\documentclass{article}

\PassOptionsToPackage{numbers}{natbib}
\usepackage[preprint]{neurips_2024}
\usepackage{neurips_2024}

\usepackage[utf8]{inputenc} 
\usepackage[T1]{fontenc}    
\usepackage{hyperref}       
\usepackage{url}            
\usepackage{booktabs}       
\usepackage{amsfonts}       
\usepackage{nicefrac}       
\usepackage{microtype}      
\usepackage{xcolor}         
\usepackage{graphicx}
\usepackage{subfig}
\usepackage{algorithm}
\usepackage{algorithmic}
\usepackage{xcolor}
\usepackage{multirow}
\usepackage{enumitem}
\usepackage{makecell}
\usepackage{amsmath}
\usepackage{tabularx}
\usepackage{capt-of}
\usepackage{wrapfig}
\usepackage{float}
\usepackage{subcaption}

\definecolor{OBred}{HTML}{DF6F63}
\definecolor{COgreen}{HTML}{6FB858}
\definecolor{INblue}{HTML}{4F8BF7}

\title{Explainable Few-shot Knowledge Tracing}

\author{%
    Haoxuan Li \\
    Beihang University\\
    \And
    Jifan Yu \\
    Tsinghua University\\
    \And
    Yuanxin Ouyang\thanks{corresponding author} \\
    Beihang University\\
    \And
    Zhuang Liu \\
    Beihang University\\
    \And
    Wenge Rong \\
    Beihang University\\
    \And
    Juanzi Li \\
    Tsinghua University\\
    \And
    Zhang Xiong \\
    Beihang University\\
    \And
    \texttt{\{lucasli, oyyx, liuzhuang, w.rong, xiongz\}@buaa.edu.cn}\\
    \texttt{\{yujifan, lijuanzi\}@tsinghua.edu.cn} \\
}

\begin{document}

\maketitle

\begin{abstract}
Knowledge tracing (KT), aiming to mine students' mastery of knowledge by their exercise records and predict their performance on future test questions, is a critical task in educational assessment. While researchers achieved tremendous success with the rapid development of deep learning techniques, current knowledge tracing tasks fall into the cracks from real-world teaching scenarios. Relying heavily on extensive student data and solely predicting numerical performances differs from the settings where teachers assess students' knowledge state from limited practices and provide explanatory feedback. To fill this gap, we explore a new task formulation: \textbf{Explainable Few-shot Knowledge Tracing}. By leveraging the powerful reasoning and generation abilities of large language models (LLMs), we then propose a cognition-guided framework that can track the student knowledge from a few student records while providing natural language explanations. Experimental results from three widely used datasets show that LLMs can perform comparable or superior to competitive deep knowledge tracing methods. We also discuss potential directions and call for future improvements in relevant topics.
\end{abstract}

\section{Introduction}

Knowledge tracing is a well-established problem originated from educational assessment~\cite{corbett1994knowledge} aiming to dynamically model students' knowledge mastery and predict their future learning performances. 
With the advancement of deep learning, models leveraging recurrent neural networks (RNNs) and attention mechanisms have gradually become mainstream for knowledge tracing~\cite{piech2015deep_DKT,ghosh2020context_AKT,choi2020towards_SAINT}. In recent years, KT research has shown two notable and promising directions. On the one hand, researchers attempt to incorporate multiple types of side information (e.g., exercise texts~\cite{liu2019ekt}, knowledge concept relationships and mappings~\cite{nakagawa2019graph_GKT,pandey2020rkt}, students' problem-solving behaviors~\cite{long2022improving_CoKT}) with students' exercise history to more accurately model their knowledge states. On the other hand, they try to reveal the links between the latent representations learned by the models and factual data to provide interpretability for the knowledge tracing models~\cite{tong2020structure_SKT,minn2022interpretable,zhu2023stable}. 

Despite the numerous attempts and decent success, the current knowledge tracing task leaves gaps in reflecting real-world scenarios where teachers evaluate students' knowledge states. On the one hand, It relies on extensive student exercise records to train deep learning knowledge tracing models to achieve remarkable performance. In contrast, in real teaching scenarios, teachers can analyze students' mastery of knowledge from a limited number of practices. On the other hand, unlike teachers can infer students' answers by analyzing and explaining their knowledge mastery, the current task is simplified to only predicting whether a student will answer the test questions correctly, mainly by deep learning sequential predictive models. The black-box nature inherent in these models hampers exploring interpretability as they represent student knowledge states as hidden vectors. Apart from the abovementioned gaps, knowledge tracing models or frameworks encounter difficulties unifying and utilizing the multi-dimensional information collected from learning environments (e.g., student behaviors, question text, knowledge relations). Primarily proposed non-generative sequential models make it challenging for such tasks to come out of numerical prediction and extend to other scenarios, such as open-ended exercising and programming learning.

\begin{figure}[t]
\centering
\includegraphics[width=\textwidth]{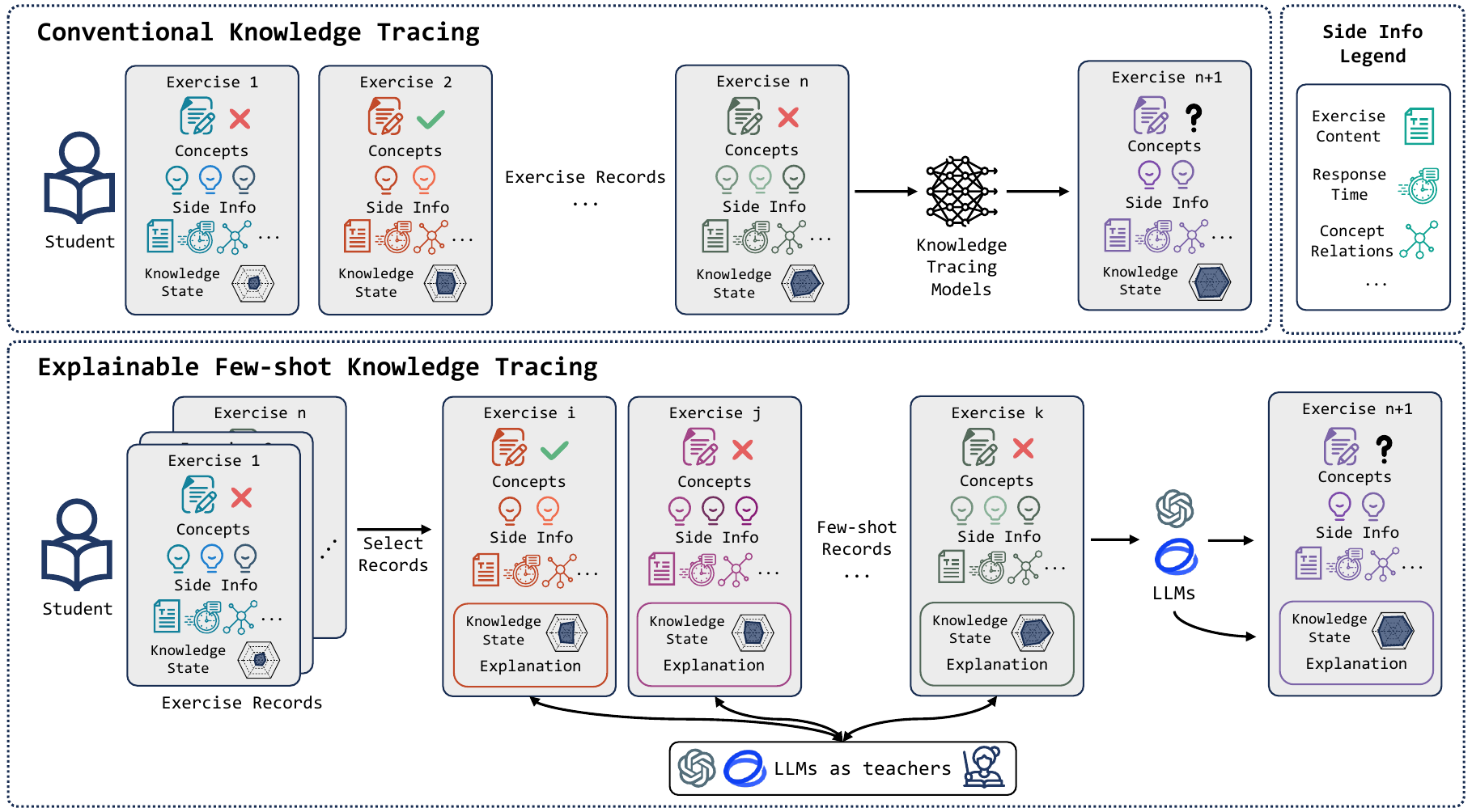}
\vspace{-3mm}
\caption{Conventional knowledge tracing and explainable few-shot knowledge tracing}
\label{introduction}
\end{figure}

In recent years, the emergence and widespread utilization of large language models (LLMs) have provided potential solutions to fill the gaps. LLMs' capability to follow complex instructions with only a few examples and provide natural language feedback makes it possible to reform the current knowledge tracing paradigm.
Inspired by the success of LLMs in other fields, we improve upon the existing knowledge tracing task formulation and propose \textbf{Explainable Few-shot Knowledge Tracin}g. As illustrated in Figure \ref{introduction}, compared to traditional knowledge tracing, explainable few-shot knowledge tracing takes a small number of informative student exercise records as input, tracks students' mastery of knowledge, and predicts future performances through reasoning while providing reasonable explanations.
Furthermore, leveraging LLMs' strong reasoning and generation abilities, the knowledge tracing task can readily adapt to diverse teaching scenarios with simple adjustments, presenting new opportunities for applying knowledge tracing in multiple educational scenarios. 
The key contributions of this paper are as follows:
\noindent
\begin{itemize}[leftmargin=*]
\item We analyze the deficiencies of conventional knowledge tracing and propose the explainable few-shot knowledge tracing task that aligns better with real teaching scenarios.
\item We introduce a cognition-guided framework that combines large language models and educational assessment principles to practice explainable few-shot knowledge tracing.
\item We adapt three public datasets and conduct experiments using open-source and closed-source LLMs. The results demonstrate that LLMs can perform comparable or superior to competitive knowledge tracing models. Furthermore, based on empirical observations from multi-perspective experiments, we suggest several potential directions for improvement.
\end{itemize}

\vspace{-3mm}

\section{Background}

\subsection{Knowledge Tracing}

\textbf{Educational Assessment and Bayesian Knowledge Tracing}
Educational assessment aims to analyze students' knowledge states, and an assessment system is generally considered to comprise three main components: observation, cognition, and interpretation~\cite{glaser2001knowing}. Cognition refers to a model of how students represent knowledge. With the introduction of the knowledge tracing concept~\cite{corbett1994knowledge}, researchers have estimated students' knowledge states by analyzing their response records. Traditional knowledge tracing (KT) methods consist of two classics: Bayesian knowledge tracing (BKT) and factor analysis models~\cite{ZhangXiong}. BKT is a hidden Markov model that treats each learner's knowledge state as a binary variable and utilizes Bayesian inference to update the state~\cite{yudelson2013individualized_BKT}. In contrast, factor analysis models aim to learn generalized parameters from historical data~\cite{yen2006item_IRT}.

\textbf{Deep Knowledge Tracing}
Recently, numerous researchers have integrated deep neural networks into KT tasks for their effectiveness and outstanding performance. Piech et al. pioneered deep learning for KT using recurrent neural networks (RNNs) to process interaction sequences over time~\cite{piech2015deep_DKT} and proposed deep knowledge tracing task setting. With the success of deep learning techniques in other domains, such as word2vec~\cite{mikolov2013efficient_word2vec} and graph neural networks~\cite{kipf2016semi_gcn, velickovic2017graph_gat}, researchers recognized the potential to leverage these techniques by incorporating auxiliary information of questions~\cite{ghosh2020context_AKT,liu2019ekt}, knowledge concepts~\cite{nakagawa2019graph_GKT,tong2020structure_SKT}, and students' learning behaviours~\cite{xu2023learning}. Moreover, attention-based models~\cite{choi2020towards_SAINT,pandey2019self_SAKT} were introduced to tackle the computational expense and instability with long sequences of RNNs. 
While achieving success in performances, the lack of interpretability raised greater attention, as the model should provide transparency and understanding of the reasoning behind learning behaviors over just the outcomes. Models incorporating educational theories like the Rasch model~\cite{ghosh2020context_AKT,bond2013applying} and the transfer of knowledge~\cite{tong2020structure_SKT} were proposed to enhance interpretability. Minn et al.~\cite{minn2022interpretable} introduced causal relationships within latent features extracted from students' behaviors. Zhu et al.~\cite{zhu2023stable} attempted to introduce causal inference for explanatory KT analysis.

Despite the remarkable success, deep knowledge tracing tasks remain a few challenges. Most methods demand extensive student exercise logs for model training, aiming to make binary predictions, which differ from the real analyzing scenarios. The black-box nature inherent in deep learning models and numerical predictions limits the explainability and struggle to generalize to other teaching scenarios, such as open-ended knowledge tracing~\cite{piech2012modeling,liu2022open} and programming learning~\cite{piech2015learning}.

\subsection{Large Language Models}

Large language models (LLMs) typically refer to transformer-based models containing hundreds of billions of parameters with multi-head attention layers stacked in very deep neural networks~\cite{zhao2023survey_llmsurvey}. LLMs can be categorized as open-sourced, like the LLaMA~\cite{touvron2023llama} and GLM~\cite{du2021glm} series, or close-sourced, like GPT-4. Trained on massive text data, LLMs exhibit solid natural language understanding to follow complex instructions and solve complex tasks due to their "emergent abilities" - capabilities not present in small language models but arising in large ones~\cite{wei2022emergent}. Additionally, LLMs can leverage multi-dimensional information for reasoning and generate natural language responses. 
Currently, researchers have achieved decent success across domains like weather forecasting~\cite{bi2022pangu}, recommendation~\cite{bao2023tallrec}, and medicine~\cite{thirunavukarasu2023large}. 
Advances in LLMs also brought new possibilities for education, where multiple aspects (e.g., teacher assistance, adaptive learning, and learning tools) benefit from the application of LLMs~\cite{ZhangXiong,Automated,li2023adapting,wang2024large}. 
The accomplishments in other fields indicate that applying LLMs to knowledge tracing could lead to similar success.

However, of the less exploration is the work of utilizing LLMs for knowledge tracing. Neshaei et al.~\cite{neshaei2024towards} explore extending the sequence modeling capabilities of LLMs to knowledge tracing. It was found that fine-tuned LLMs outperformed naive baselines and matched Bayesian knowledge tracing, suggesting further refinements and a deeper understanding of their predictive mechanisms could enhance performance. Despite the first attempt, it is limited by the original knowledge tracing task settings, where LLMs cannot handle such extensive student exercise records. It motivates us to explore a knowledge tracing paradigm for the era of large language models, and to leverage the advantages of LLMs to address the shortcomings of traditional knowledge tracing settings.

\section{Explainable Few-shot Knowledge Tracing}

Deep knowledge tracing task~\cite{piech2015deep_DKT} is formulated as estimating student next state $\hat{s}_{t+1}$ given student records $\mathcal{X}_t$, questions $\mathcal{Q}$, knowledge concepts $\mathcal{C}$ and a knowledge tracing model $M_{DP}$, denoted as,
\begin{align}
    \mathcal{X}_t&=\{x_0,...,x_t\}, \mathcal{F}: \mathcal{Q} \rightarrow \mathcal{C}, \\
    \hat{s}_{t+1}&=\arg \max_y P(y \mid M_{DP}, \mathcal{X}_t, \mathcal{Q}, \mathcal{C}),
    \label{dkt def}
\end{align}
where $\mathcal{F}$ denotes the mapping relations of exercises and knowledge concepts.
We then define the explainable few-shot knowledge tracing by integrating selected student records $\mathcal{X}_t^\prime \in \mathcal{X}_t$, preditive model $M$, questions $\mathcal{Q}^\prime$ and knowledge concepts $\mathcal{C}^\prime$ with extended information to output estimated student states $\hat{s}_{t+1}^\prime$ and generate explanation $\hat{E}$, further formulated as,
\begin{align}
    \hat{s}_{t+1}^\prime, \hat{E}&=\arg \max_{\omega,\phi} P(\omega,\phi \mid \mathcal{M}, \mathcal{X}_t^\prime, \mathcal{Q}^\prime, \mathcal{C}^\prime).
    \label{ex-fs-kt def}
\end{align}

\subsection{Overview}
\begin{figure}[t]
\centering
\includegraphics[width=\textwidth]{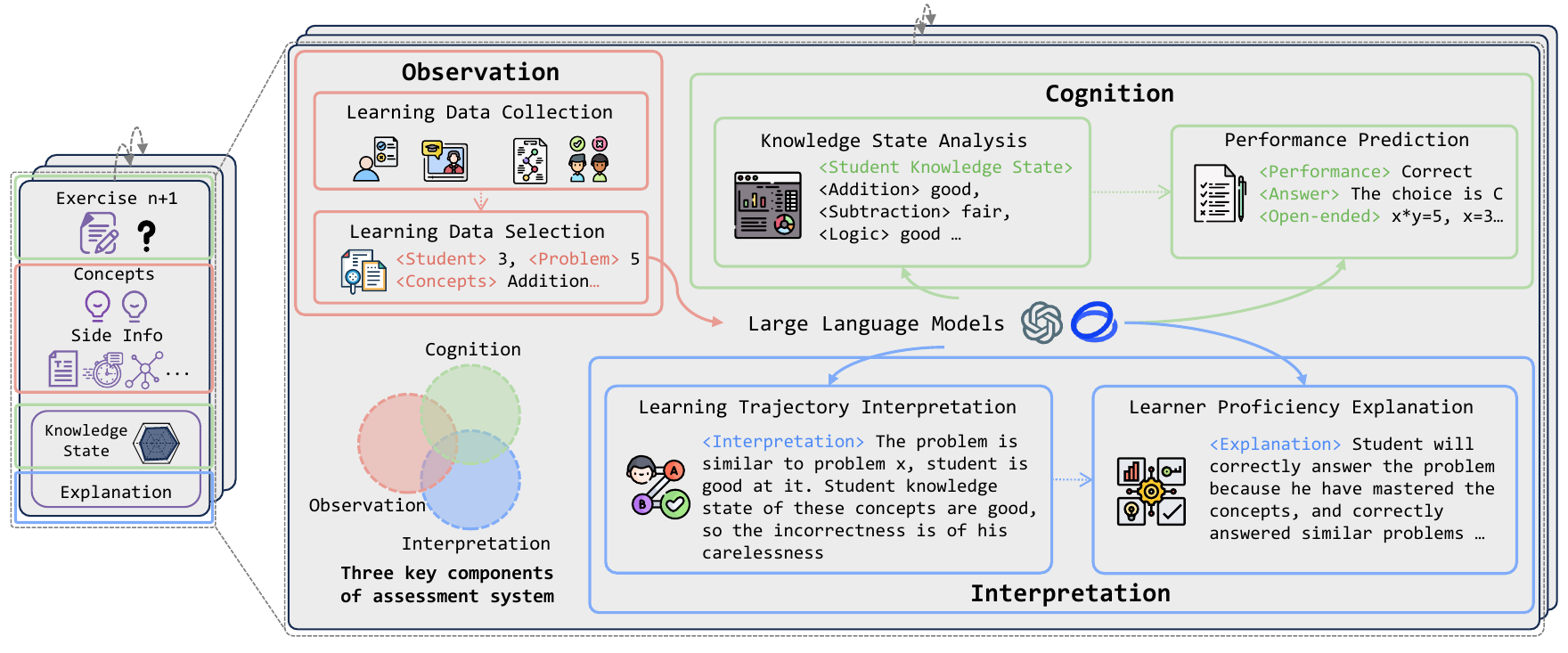}
\caption{The cognition-guided framework for explainable few-shot knowledge tracing}
\label{framework}
\end{figure}

To further practice explainable few-shot knowledge tracing tasks, we propose a cognition-guided framework by leveraging large language models, which consists of three indispensable fundamental components~\cite{glaser2001knowing} originated from assessment systems: \textbf{Observation}, denoted as $M_O$, defining the task scenario and collecting data, \textbf{Cognition}, represented as $M_C$, which models learners' knowledge state and predict performances, and \textbf{Interpretation}, denoted as $M_I$, providing explanations of assessing processes. 
Notably, most existing knowledge tracing models primarily function as the Cognition module, while a few explainable knowledge tracing methods partially assume the role of the Interpretation module. As depicted in Figure \ref{framework}, by leveraging large language models, we can unify the Cognition and Interpretation, enabling them to ingest the diverse data acquired by the Observation module, integrating the three components to form a cohesive system. We further dive into the three components and then clarify how they work together within the new task.

\subsection{Observation}

\textbf{Observation}, \textit{which is the task or situation that allows one to observe students' performance}~\cite{glaser2001knowing}, defines the learning environment within the assessment system, determining factors such as the type of knowledge acquired by the learner and the tasks they engage with. It collects multi-dimensional and multi-modal data $\mathcal{X}_{raw}$ from the designated learning environment $\mathcal{E}$, generating the necessary inputs for subsequent assessment processes, and thus comprises two sub-modules: Learning Data Collection $M_{dc}$ and Learning Data Selection $M_{ds}$.

\textbf{Learning Data Collection} determines the data types to be collected based on the task scenario and carrying out the collection process. Typically, it involves gathering information such as the student's response sequence, including correctness, timestamps, and duration, as well as question-related information like problem contents and knowledge concepts, and forms structured dataset $\mathcal{X}_c$.

\textbf{Learning Data Selection} curates the processed data $\mathcal{X}_c$ through strategic selection and reorganizing, providing the refined inputs $\mathcal{X}_s = \{x_1,...,x_s\}$ to the cognition and interpretation modules as required. In the implementation, we select several exercise records to generate informative few-shots for LLMs to predict performance. For simplicity, we implement random and time-ordered linear sampling strategies to select candidate few-shots from student history exercise records.
Deep knowledge tracing models often require vast amounts of student response data for accurate prediction. In contrast, educators can frequently gauge a learner's knowledge level from a limited yet information-rich set of response records corresponding to $\mathcal{X}_s$. The advantage of large language models lies in their ability to leverage in-context learning and reasoning, enabling them to extract high-quality insights while seamlessly ingesting diverse inputs. It aligns with real-world instructional scenarios and lays the foundation for fully exploiting the strengths of LLMs for cognition and interpretation.

\subsection{Cognition}

\textbf{Cognition}, \textit{which is a model of how students represent knowledge \& develop competence in the domain}~\cite{glaser2001knowing}, synthesizes a comprehensive representation of the learner's evolving knowledge state $\hat{K}_s$ and generate predictions $\hat{P}$ from $\mathcal{X}_s$. This module is divided into two sub-modules: Knowledge State Analysis $M_{ca}$ and Performance Prediction $M_{cp}$.

\textbf{Knowledge State Analysis} dynamically analyzes the learner's mastery of knowledge throughout the practice process by $\mathcal{X}_s$, containing student response records, question information, and behavioral patterns. It generates reliable knowledge state estimates $\hat{K}_s$ as essential references for performance prediction and Interpretation, formulated as,
\begin{align}
    \label{equ:KSA}
        \hat{k}_j & = M_{ca}(\mathcal{X}_j, \hat{K}_{j-1}, \hat{I}_{j-1}),
        \\
        M_{ca}(\cdot) & =\arg\max_{\omega} P(\omega \mid \cdot, \omega_{c_aprompts}),
\end{align}
where $\hat{k}_j$ is the estimated knowledge state with respect to $x_j$ and $\hat{K}_{j-1} = \{\hat{k}_1, \hat{k}_2, \ldots, \hat{k}_{j-1}\}$. $\omega_{c_aprompts}$ is the prompts designed for knowledge state analysis, and $\hat{I}_{j-1}$ is the set of interpretation of $\hat{k}_1$ to $\hat{k}_{j-1}$, which is elaborated in section \ref{subsection:Interpretation}.
In the implementation, LLMs are asked to generate student mastery of knowledge with ternary value (good, fair, or fail) for each concept contained in the exercise the student encounters. It is also worthwhile to explore other customized analysis, deriving the benefits from the flexibility and generalizability of large language models compared to functionally similar cognitive diagnosis models.

\textbf{Performance Prediction} forecasts the learner's performance $\hat{P}$ on predefined environment $\mathcal{E}$ by mining selected data $\mathcal{X}_s$, estimated state $\hat{K}_s$ and interpretation $\hat{I}$ from $M_I$, denoted as,
\begin{align}
    \label{equ:PP}
        \hat{P} &= M_{cp}(\mathcal{X}_s, \hat{K}_{s}, \hat{I}_{s}, x_p),
        \\
        M_{cp}(\cdot)&=\arg\max_{\omega} P(\omega \mid \cdot, \omega_{c_pprompts}).
\end{align}
$x_p$ is the data of exercise to predict and $\omega_{c_pprompts}$ is the prompts designed for predicting performance.Traditionally, performance is quantified as the probability of correctness or percentage scores. However, by leveraging the generative capabilities of large language models, we can extend the prediction to a broader range of learning scenarios less explored by deep learning models, such as open-ended question answering and programming tasks. 

\subsection{Interpretation}
\label{subsection:Interpretation}

\textbf{Interpretation}, \textit{which is a method for making sense of the data relative to our cognitive model}~\cite{glaser2001knowing}, leverages $\hat{P}, \hat{K}_s$ from the previous modules to generate diagnostic feedback and interpretable analytical insights. These insights facilitate targeted pedagogical interventions to optimize the learner's educational experience and provide a mechanism to evaluate and justify the validity of the observation module's task design and data selection strategies. The interpretation module comprises two sub-modules: Learning Trajectory Interpretation $M_{it}$ and Learner Proficiency Explanation $M_{pe}$.

\textbf{Learning Trajectory Interpretation} harnesses data $\mathcal{X}_s$, and the knowledge estimates $\hat{K}_s$ to furnish natural language explanations $\hat{I}_s$ for the learner's historical practice behaviors, formulated as,
\begin{align}
    \label{equ:LTI}
        \hat{i}_j & = M_{it}(\mathcal{X}_j, \hat{K}_{j}, \hat{I}_{j-1}),
        \\
        M_{it}(\cdot)&=\arg\max_{\omega} P(\omega \mid \cdot, \omega_{i_tprompts}),
\end{align}
where $\hat{i}_j$ is the interpretation of student records $x_j$ and $\hat{I}_{j-1} = \{\hat{i}_1, \hat{i}_2, \ldots, \hat{i}_{j-1}\}$. For instance, if a learner exhibits proficiency in the concepts related to a question but still provides an incorrect answer, it may attribute the error to carelessness, offering a plausible explanation. Importantly, these explanations can inform and refine the knowledge state analysis and performance prediction within Cognition, accounting for transient factors without unduly penalizing the learner's estimated knowledge state. In contrast, conventional deep learning models may inaccurately degrade the learner's knowledge states due to occasional carelessness, resulting in erroneous predictions. The versatility of large language models enables us to encompass and interpret a wide array of learner behaviors, furnishing more reliable and effective natural language explanations than the numerical interpretations provided by existing explainable knowledge tracing models.

\textbf{Learner Proficiency Explanation} integrates $\mathcal{X}_s$ from observation, the estimated knowledge state $\hat{K}_s$, and the explanations $\hat{I}_s$ to provide meaningful insights $\hat{E}$ into the performance predictions of the Cognition module. It clarifies the complex interplay between learners' proficiency levels, learning habits, and task performance by situating these predictions within specific instructional scenarios and learner task contexts. The final process can be formulated as,
\begin{align}
\label{equ:LPE}
        \hat{E} &= M_{pe}(\mathcal{X}_s, \hat{K}_{s}, \hat{I}_{s}, x_p, \hat{P}),
        \\
        M_{pe}(\cdot)&=\arg\max_{\omega} P(\omega \mid \cdot, \omega_{p_eprompts}).
\end{align}
$\hat{E}$ promotes a nuanced explanation of learners' competencies, empowering educators to make timely adjustments to teaching content and cater to individual needs.

\vspace{-1.5mm}
\section{Experiments}
\vspace{-1.5mm}
In this section, we detail the approach to practicing explainable few-shot knowledge tracing, including the construction of datasets and model implementation. We compare the performance of LLMs on this task against deep learning models on traditional knowledge tracing. Furthermore, we investigate potential improvement when employing LLMs for this task.

\vspace{-1.5mm}
\subsection{Task Setups}
\label{setups}
\vspace{-1.5mm}

\begin{wrapfigure}{r}{0.4\textwidth}
    \centering
    \vspace{-5mm}
    \begin{center}
    \includegraphics[width=0.35\textwidth]{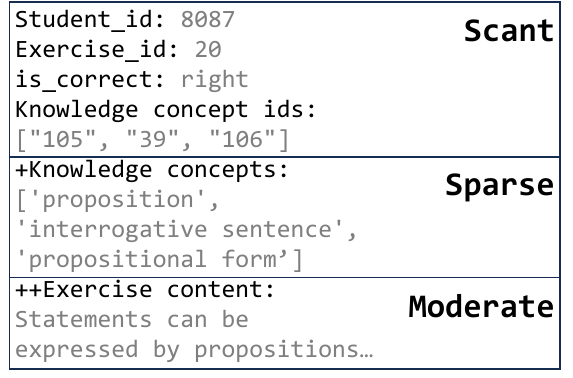}
    \caption{Dataset of different modes.}
    \label{fig:dataset_mode}
    \end{center}
    \vspace{-5mm}
\end{wrapfigure}

\textbf{Datasets} We selected three public datasets: FrcSub\footnote{http://staff.ustc.edu.cn/\%7Eqiliuql/data/math2015.rar}, MOOCRadar~\cite{yu2023moocradar}, and XES3G5M~\cite{liu2024xes3g5m}.
The detailed statistics of these three datasets are presented in Appendix \ref{appendix:datasets}.
The task can integrate multidimensional information as input by designing structured textual data and appropriate prompts. 
Depending on the type of side information incorporated, we created different modes:scant and sparse for three datasets, and additional moderate mode for MOOCRadar and XES3G5M, varying degrees of information richness, shown in Figure \ref{fig:dataset_mode}. The scant mode utilizes only the primary student ID, exercise ID, skill ID, and student interaction records. The sparse mode builds upon scant by incorporating skill representation information. The moderate further includes textual descriptions of exercises over the sparse. 

\textbf{Metrics}
We collect accuracy, precision, recall, and F1 scores as evaluation metrics, as the area under the curve (AUC) cannot be employed since LLMs provide binary predictions. Due to page constraints, the experimental results of precision and recall metrics will be included in the Appendix \ref{appendix:overall performance} and \ref{appendix:more_pres_recalls}.

\textbf{Baselines}
We select several commonly employed and competitive baselines in knowledge tracing:
1) \textbf{DKT}~\cite{piech2015deep_DKT} employs LSTM layers to encode the students' knowledge state and predict their performance on exercises.
2) \textbf{DKVMN}~\cite{zhang2017dynamic_DKVMN} designs a static key matrix to capture relations between knowledge components and a dynamic value matrix to track the evolution of students' knowledge.
3) \textbf{GKT}~\cite{nakagawa2019graph_GKT} leverages graph structure to model interactions between exercises.
4) \textbf{AKT}~\cite{ghosh2020context_AKT} utilizes an attention mechanism to characterize the temporal distance between questions and the student's history interactions.
5) \textbf{SAKT}~\cite{pandey2019self_SAKT} incorporates a self-attention module to capture latent relations between exercises and student responses.
6) \textbf{SAINT}~\cite{choi2020towards_SAINT} adopts a transformer architecture to jointly model sequences of exercises and responses.

\vspace{-1.5mm}
\subsection{Overall Performance}
\vspace{-1.5mm}

We compared the best performance of GLM3-6B~\cite{du2021glm}, GLM4, and GPT-4 across all modes of three datasets with considered baselines.  Notably, all considered baselines requires the full training set to achieve best performances, whereas ours only require a few, and such a small amount is far from enough for the baselines. The best three metrics in each column are marked using \textbf{bold}, \underline{underlined}, and \textit{italics}. Overall, GLM4 and GPT-4 performed comparable or superior to the baselines on all three datasets. Notably, on the MOOCRadar dataset, GLM4 and GPT-4 outperformed all baselines, showing improvements of 3.01\% and 1.66\% in Accuracy and F1 Score, respectively. It demonstrates that leveraging LLMs within explainable few-shot knowledge tracing can match or surpass conventional deep learning models. In contrast, GLM3-6B did not perform as well as expected, which could be attributed to the extensive input context. During experiments, we observed that the GLM3-6B often struggled to follow instructions, indicating that a fine-tuned small model may potentially achieve better performances. Specifically, we present more comprehensive results in Appendix \ref{appendix:overall performance}, and the implementation details to achieve the best performance in Appendix \ref{appendix:Reproducibility}.

\begin{table*}[t]
	\centering
	\caption{A comparison of the accuracy and F1 score among baselines in three datasets.}

        \begin{tabular}{c|l| c c| c c| c c}
            \toprule
            \multirow{3}{*}{\textbf{Input scale}}&\multirow{3}{*}{\textbf{Baselines}} & \multicolumn{6}{c}{\textbf{Dataset}} \\
            \cline{3-8}
             & & \multicolumn{2}{c|}{FrcSub}
            & \multicolumn{2}{c|}{MOOCRadar}
            & \multicolumn{2}{c}{XES3G5M} \\
            \cline{3-8}
             \multicolumn{1}{c|}{} & \multicolumn{1}{c|}{} & \multicolumn{1}{c}{Accuracy} & \multicolumn{1}{c|}{F1 Score} & \multicolumn{1}{c}{Accuracy} & \multicolumn{1}{c|}{F1 Score} & \multicolumn{1}{c}{Accuracy} & \multicolumn{1}{c}{F1 Score} \\
            \midrule
                \multirow{6}{*}{full-set} & DKT & 0.7481 & 0.7514 & 0.8210 & 0.8882 & 0.8355 & 0.9017 \\
                 & DKVMN & 0.7909 & \underline{0.8077} & 0.8147 & 0.8836 & 0.8372 & \textit{0.9037} \\
                 & GKT* & 0.5480 & 0.3043 & 0.7991 & 0.8772 & 0.8169 & 0.8923 \\
                 & AKT & 0.7747 & 0.7869 & 0.8194 & 0.8870 & \underline{0.8435} & \textbf{0.9063} \\
                 & SAKT & 0.7476 & 0.7389 & 0.7956 & 0.8706 & 0.8298 & 0.8990 \\
                 & SAINT & \textbf{0.8061} & \textbf{0.8201} & \textit{0.8241} & \textit{0.8904} & \textit{0.8399} & \underline{0.9044} \\
                \hline
                \multirow{3}{*}{few-shots} & GLM3-6b & 0.6571 & 0.6496 & 0.5378 & 0.6753 & 0.5434 & 0.6580 \\
                 & GLM4 & \textit{0.7939} & \textit{0.7889} & \textbf{0.8489} & \textbf{0.9052} & \textbf{0.8491} & 0.8978 \\
                 & GPT-4 & \underline{0.7968} & 0.7471 & \underline{0.8246} & \underline{0.9029} & 0.8176 & 0.8714 \\
            \bottomrule
        \end{tabular}
	\label{overall performance}
\vspace{-3mm}
\end{table*}

\vspace{-2mm}
\begin{figure}[b]
\centering
\includegraphics[width=\textwidth]{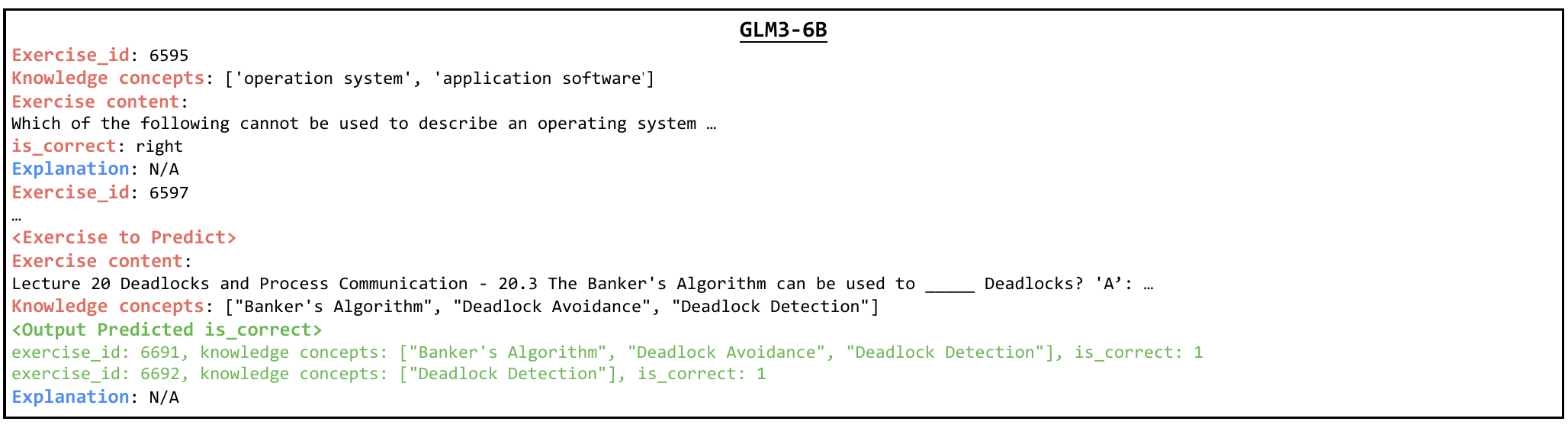}
\caption{Case study of GLM3-6B}
\label{case study_2}
\end{figure}
\vspace{-2mm}

\begin{figure}[t]
\centering
\includegraphics[width=\textwidth]{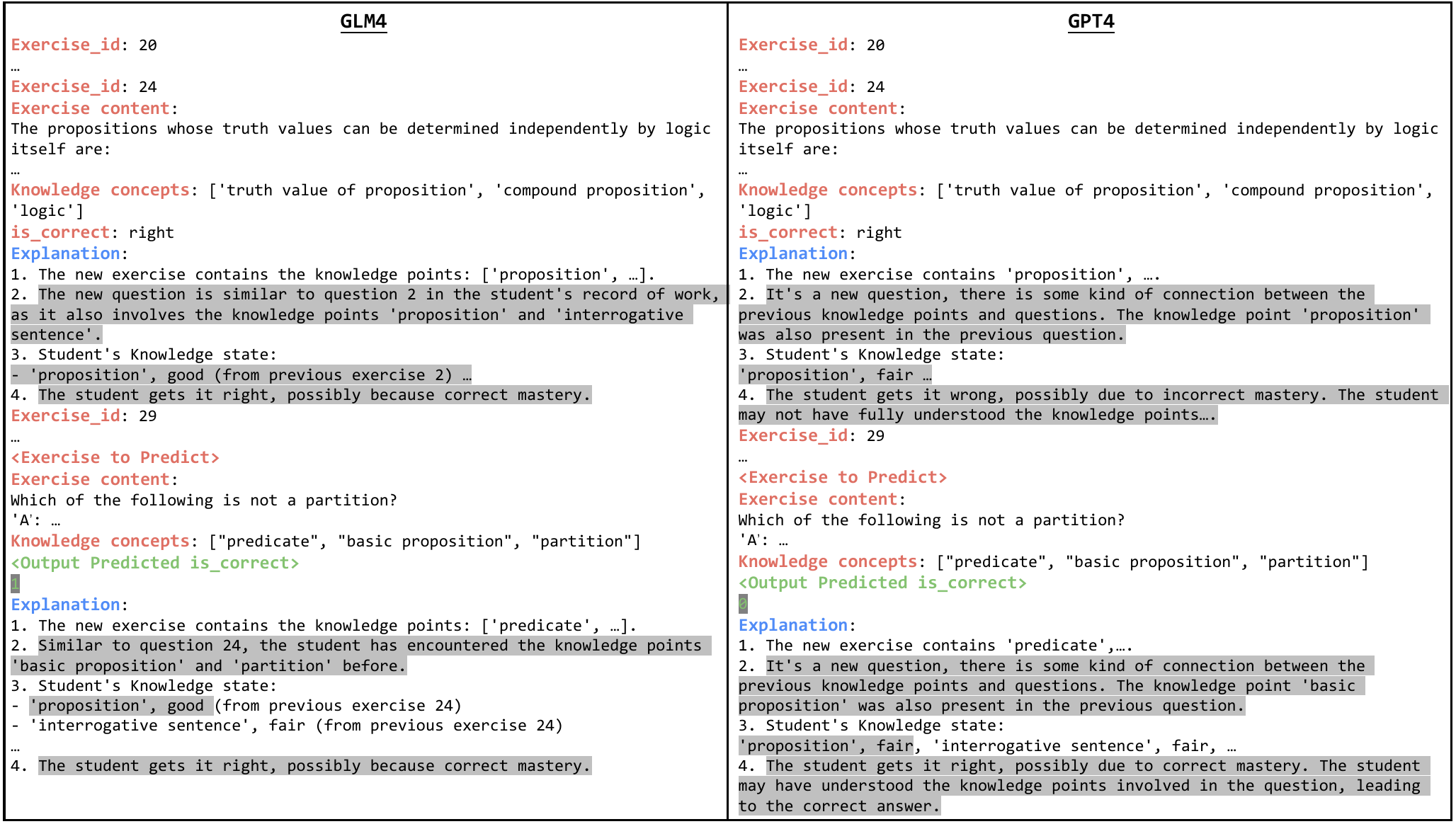}
\vspace{-3mm}
\caption{Case study of GLM4 and GPT-4}
\label{case study_1}
\end{figure}

\vspace{-2mm}
\subsection{Case Study}
\vspace{-1.5mm}
We randomly select examples of all considered LLMs from the MOOCRadar-moderate. It involves estimating the student's knowledge state in student history records, predicting student performances, and providing an explanation. Identifiers are colored to correspond with modules in Figure \ref{framework}.
The content before <Exercise to Predict> contained the previous context, including four few-shots and the LLM's analysis of the student's responses based on the selected examples. <Exercise to Predict> contains the information of the test exercise. <Output Predicted is\_correct> represented the LLM's prediction of the student's performance on <Exercise to Predict>, followed by an explanation for the prediction. For detailed prompts and more cases, please refer to Appendices \ref{appendix:prompts} and \ref{appendix:more cases}.

\textbf{GLM3-6B}
As shown in Figure \ref{case study_1}, we removed the knowledge state analysis for each student exercise record to limit the input length for GLM3-6B. However, in many cases, even though it output predictions on all exercises the student had encountered, it failed to satisfactorily meet our required output format, which is only one 0 or 1 for a single test exercise.

\textbf{GLM4 \& GPT-4} 
Illustrated in Figure \ref{case study_1}, we highlighted the differences between the outputs of GLM4 and GPT-4 in gray. We observed that both models are able to follow the instructions and generate formatted explanations in most cases. Differences occurs where GPT-4 incorrectly assumed the student had answered Exercise 24 incorrectly and provided an explanation for this assumption. Furthermore, when explaining its prediction, GPT-4 failed to recognize the incorrectness of the student's performance and instead offered an explanation suggesting the student had answered correctly. This issue was also present in some cases for GLM4, possibly due to the models' limited context window to accurately identify such shot and specific information.

\subsection{Discussion}
\label{exp:discussion}
We will discuss empirical observations from designed experiments and the potential directions for improving performances when leveraging LLMs for explainable few-shot knowledge tracing.

\begin{table*}[b]
	\centering
	\caption{Performance comparison of different number of few-shots of GLM4 on three datasets.}
        \begin{tabular}{l| l l| l l| l l}
            \toprule
            \multirow{2}{*}{GLM4} & \multicolumn{2}{c|}{\textbf{FrcSub-sparse}} & \multicolumn{2}{c|}{\textbf{XES3G5M-sparse}} & \multicolumn{2}{c}{\textbf{XES3G5M-moderate}} \\
            \multicolumn{1}{c|}{} & \multicolumn{1}{c}{Accuracy} & \multicolumn{1}{c|}{F1 Score} & \multicolumn{1}{c}{Accuracy} & \multicolumn{1}{c|}{F1 Score} & \multicolumn{1}{c}{Accuracy} & \multicolumn{1}{c}{F1 Score} \\
            \midrule
                  4 & 0.7192$^{\textcolor{gray}{+0.0}}$ & 0.7086$^{\textcolor{gray}{+0.0}}$ & 0.4399$^{\textcolor{gray}{+0.0}}$ & 0.4707$^{\textcolor{gray}{+0.0}}$ & 0.6672$^{\textcolor{gray}{+0.0}}$ & 0.7592$^{\textcolor{gray}{+0.0}}$ \\
                 
                  8 & 0.7771$^{\textcolor{red}{+8.1}}$ & 0.7568$^{\textcolor{red}{+6.8}}$ & 0.7057$^{\textcolor{red}{+60.4}}$ & 0.7752$^{\textcolor{red}{+64.7}}$ & 0.7928$^{\textcolor{red}{+18.8}}$ & 0.8623$^{\textcolor{red}{+13.6}}$ \\
                 
                  16 & 0.7939$^{\textcolor{red}{+10.4}}$ & 0.7889$^{\textcolor{red}{+11.3}}$ & 0.7542$^{\textcolor{red}{+71.4}}$ & 0.8395$^{\textcolor{red}{+78.4}}$ & 0.8491$^{\textcolor{red}{+27.3}}$ & 0.8978$^{\textcolor{red}{+18.3}}$ \\
            \bottomrule
        \end{tabular}
	\label{number_of_fs_comparison}
        \vspace{-3mm}
\end{table*}

\begin{figure}[t]
\centering
\includegraphics[width=\textwidth]{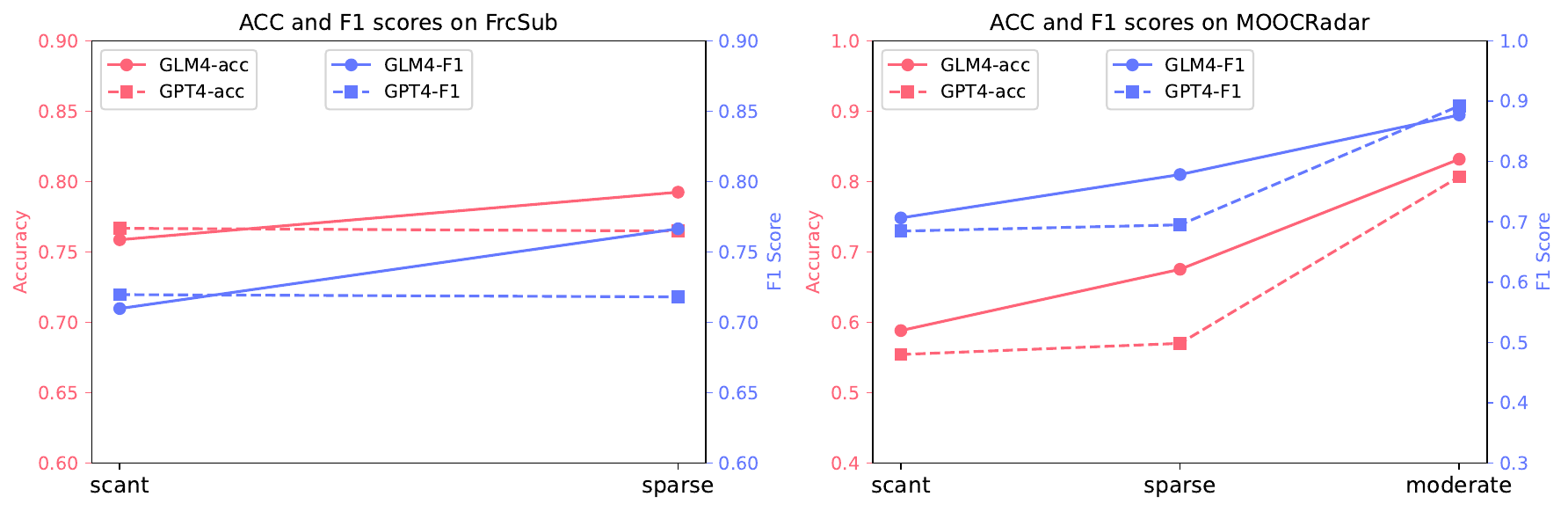}
\vspace{-3mm}
\caption{Performance of GLM4 and GPT-4 on FrcSub and MOOCRadar}
\label{information_richness}
\end{figure}

\textbf{Exercise text helps a lot; knowledge concepts do a little.}
We analyze the performance of GLM4 and GPT-4 on the different modes in FrcSub dataset and MOOCRadar dataset, as shown in Figure \ref{information_richness}. "GLM4-acc" denotes the accuracy metrics for GLM4 selecting first 4 few-shots.
It can be observed that the performances substantially improve from sparse to moderate mode in MOOCRadar. Integrating only knowledge concepts gains a relatively lower improvement or even a slight decline in Frcsub using GPT-4. It indicates that combining exercise textual information benefits more than knowledge concepts since exercise texts provide more contexts, and concepts provide less than those using IDs.
Therefore, a key consideration for boosting performance lies in fully leveraging the existing datasets, formulating them into structured texts, and designing proper prompts that enable LLMs to utilize the additional information effectively.

\textbf{Increasing the number of few-shots benefits, but too much leads to confusion.}
We analyze GLM4's performance when using 4, 8, and 16 randomly selected few-shots to explore the impact of different numbers of few-shots on the final results. 
As shown in Table \ref{number_of_fs_comparison}, increasing the number of few-shots leads to improved performance. Notably, for the XES3G5M-sparse dataset, the accuracy saw a significant 71.4\% improvement from 0.4399 with four shots to 0.7542 with 16 shots, and the F1 score achieved an impressive 78.4\% enhancement. These results highlight the substantial benefits of utilizing more few-shots, especially for student with long records, which is presented in Appendix \ref{appendix:studentlength}.
However, excessive few-shots would result in an overly long and repeated context, hampering the LLMs' capabilities. Even with 4 few-shots, for those that are relatively small, like GLM3-6B in Figure \ref{case study_2}, it fails to follow the instructions, and for GLM4 and GPT-4 in Figure \ref{case study_1}, it leads to incorrectly capturing the student behavior information. As a consequences, it may result in generating misguided information. Therefore, developing effective memory modules enabling LLMs to leverage more few-shots for tracking students' states remains an important direction to explore.

\begin{figure}[h]
\centering
\includegraphics[width=\textwidth]{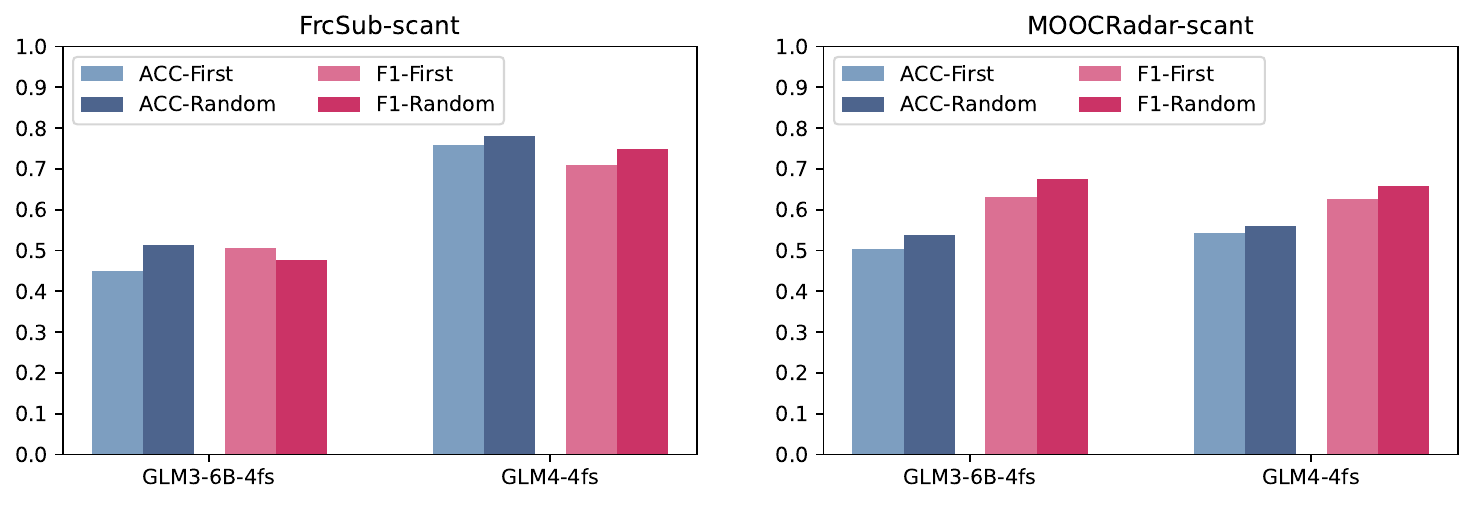}
\caption{Performance comparison of different few-shots selection strategies}
\label{selection_strategies}
\end{figure}

\textbf{Random few-shots work better in long sequences.} We investigate the impact of different few-shot selection strategies on the final performance. Figure \ref{selection_strategies} shows the performance of the "First" or "Random" selection strategies, using GLM3-6B and GLM4 on FrcSub-scant and MOOCRadar-scant datasets. Generally, the random selection outperforms selecting the first few exercises as few-shots. It is more pronounced in datasets with longer student interaction records. When student learning histories are extensive, the test exercises are more likely unrelated to the initial questions, as demonstrated in Appendix \ref{appendix:studentlength}. It is worth noting that there remains significant room for improvement in selection strategies. We recommend exploring more optimal selection methods. For instance, one could select the most recent exercise records, similar exercises to the predicted ones, or utilize retrieval-augmented generation to construct informative few-shots.

\vspace{-3mm}
\vspace{-3mm}

\section{Conclusions}
\vspace{-2mm}
We formulated the explainable few-shot knowledge tracing task to fill the gap between the conventional knowledge tracing task and real teaching scenarios and proposed a cognition-guided framework to conduct this task. We further demonstrate that LLMs can achieve comparable or superior performances to competitive baselines in conventional knowledge tracing while providing more natural language explanations under our proposed framework. Then, we discuss potential directions for further enhancing LLM performance on this task, including providing more informative relevant few-shots. The ability of large language models enables understanding student essays or programming codes, even for multi-modal inputs (e.g., drawings, speech).
By modifying the prompts of modules in the framework and incorporating specific information, it is worthwhile to extend explainable few-shot knowledge tracing to new tasks, where of the less exploration by existing methods are tasks like open-ended question answering and programming knowledge tracing.

\newpage
\bibliographystyle{neurips}
\bibliography{neurips_2024}

\begin{thebibliography}{10}

\bibitem{corbett1994knowledge}
Corbett, A.~T., J.~R. Anderson.
\newblock Knowledge tracing: Modeling the acquisition of procedural knowledge.
\newblock \emph{User modeling and user-adapted interaction}, 4:253--278, 1994.

\bibitem{piech2015deep_DKT}
Piech, C., J.~Bassen, J.~Huang, et~al.
\newblock Deep knowledge tracing.
\newblock \emph{Advances in neural information processing systems}, 28, 2015.

\bibitem{ghosh2020context_AKT}
Ghosh, A., N.~Heffernan, A.~S. Lan.
\newblock Context-aware attentive knowledge tracing.
\newblock In \emph{Proceedings of the 26th ACM SIGKDD international conference on knowledge discovery \& data mining}, pages 2330--2339. 2020.

\bibitem{choi2020towards_SAINT}
Choi, Y., Y.~Lee, J.~Cho, et~al.
\newblock Towards an appropriate query, key, and value computation for knowledge tracing.
\newblock In \emph{Proceedings of the seventh ACM conference on learning@ scale}, pages 341--344. 2020.

\bibitem{liu2019ekt}
Liu, Q., Z.~Huang, Y.~Yin, et~al.
\newblock Ekt: Exercise-aware knowledge tracing for student performance prediction.
\newblock \emph{IEEE Transactions on Knowledge and Data Engineering}, 33(1):100--115, 2019.

\bibitem{nakagawa2019graph_GKT}
Nakagawa, H., Y.~Iwasawa, Y.~Matsuo.
\newblock Graph-based knowledge tracing: modeling student proficiency using graph neural network.
\newblock In \emph{IEEE/WIC/ACM International Conference on Web Intelligence}, pages 156--163. 2019.

\bibitem{pandey2020rkt}
Pandey, S., J.~Srivastava.
\newblock Rkt: relation-aware self-attention for knowledge tracing.
\newblock In \emph{Proceedings of the 29th ACM International Conference on Information \& Knowledge Management}, pages 1205--1214. 2020.

\bibitem{long2022improving_CoKT}
Long, T., J.~Qin, J.~Shen, et~al.
\newblock Improving knowledge tracing with collaborative information.
\newblock In \emph{Proceedings of the fifteenth ACM international conference on web search and data mining}, pages 599--607. 2022.

\bibitem{tong2020structure_SKT}
Tong, S., Q.~Liu, W.~Huang, et~al.
\newblock Structure-based knowledge tracing: An influence propagation view.
\newblock In \emph{2020 IEEE international conference on data mining (ICDM)}, pages 541--550. IEEE, 2020.

\bibitem{minn2022interpretable}
Minn, S., J.-J. Vie, K.~Takeuchi, et~al.
\newblock Interpretable knowledge tracing: Simple and efficient student modeling with causal relations.
\newblock In \emph{Proceedings of the AAAI conference on artificial intelligence}, vol.~36, pages 12810--12818. 2022.

\bibitem{zhu2023stable}
Zhu, J., X.~Ma, C.~Huang.
\newblock Stable knowledge tracing using causal inference.
\newblock \emph{IEEE Transactions on Learning Technologies}, 2023.

\bibitem{glaser2001knowing}
Glaser, R., N.~Chudowsky, J.~W. Pellegrino.
\newblock \emph{Knowing what students know: The science and design of educational assessment}.
\newblock National Academies Press, 2001.

\bibitem{ZhangXiong}
Xiong, Z., H.~Li, Z.~Liu, et~al.
\newblock A review of data mining in personalized education: Current trends and future prospects.
\newblock \emph{Frontiers of Digital Education}, 2024.

\bibitem{yudelson2013individualized_BKT}
Yudelson, M.~V., K.~R. Koedinger, G.~J. Gordon.
\newblock Individualized bayesian knowledge tracing models.
\newblock In \emph{Artificial Intelligence in Education: 16th International Conference, AIED 2013, Memphis, TN, USA, July 9-13, 2013. Proceedings 16}, pages 171--180. Springer, 2013.

\bibitem{yen2006item_IRT}
Yen, W.~M., A.~R. Fitzpatrick.
\newblock Item response theory.
\newblock \emph{Educational measurement}, 4:111--153, 2006.

\bibitem{mikolov2013efficient_word2vec}
Mikolov, T., K.~Chen, G.~Corrado, et~al.
\newblock Efficient estimation of word representations in vector space.
\newblock \emph{arXiv preprint arXiv:1301.3781}, 2013.

\bibitem{kipf2016semi_gcn}
Kipf, T.~N., M.~Welling.
\newblock Semi-supervised classification with graph convolutional networks.
\newblock \emph{arXiv preprint arXiv:1609.02907}, 2016.

\bibitem{velickovic2017graph_gat}
Velickovic, P., G.~Cucurull, A.~Casanova, et~al.
\newblock Graph attention networks.
\newblock \emph{stat}, 1050(20):10--48550, 2017.

\bibitem{xu2023learning}
Xu, B., Z.~Huang, J.~Liu, et~al.
\newblock Learning behavior-oriented knowledge tracing.
\newblock In \emph{Proceedings of the 29th ACM SIGKDD conference on knowledge discovery and data mining}, pages 2789--2800. 2023.

\bibitem{pandey2019self_SAKT}
Pandey, S., G.~Karypis.
\newblock A self-attentive model for knowledge tracing.
\newblock \emph{arXiv preprint arXiv:1907.06837}, 2019.

\bibitem{bond2013applying}
Bond, T.~G., C.~M. Fox.
\newblock \emph{Applying the Rasch model: Fundamental measurement in the human sciences}.
\newblock Psychology Press, 2013.

\bibitem{piech2012modeling}
Piech, C., M.~Sahami, D.~Koller, et~al.
\newblock Modeling how students learn to program.
\newblock In \emph{Proceedings of the 43rd ACM technical symposium on Computer Science Education}, pages 153--160. 2012.

\bibitem{liu2022open}
Liu, N., Z.~Wang, R.~Baraniuk, et~al.
\newblock Open-ended knowledge tracing for computer science education.
\newblock In \emph{Proceedings of the 2022 Conference on Empirical Methods in Natural Language Processing}. 2022.

\bibitem{piech2015learning}
Piech, C., J.~Huang, A.~Nguyen, et~al.
\newblock Learning program embeddings to propagate feedback on student code.
\newblock In \emph{International conference on machine Learning}, pages 1093--1102. PMLR, 2015.

\bibitem{zhao2023survey_llmsurvey}
Zhao, W.~X., K.~Zhou, J.~Li, et~al.
\newblock A survey of large language models.
\newblock \emph{arXiv preprint arXiv:2303.18223}, 2023.

\bibitem{touvron2023llama}
Touvron, H., T.~Lavril, G.~Izacard, et~al.
\newblock Llama: Open and efficient foundation language models.
\newblock \emph{arXiv preprint arXiv:2302.13971}, 2023.

\bibitem{du2021glm}
Du, Z., Y.~Qian, X.~Liu, et~al.
\newblock Glm: General language model pretraining with autoregressive blank infilling.
\newblock \emph{arXiv preprint arXiv:2103.10360}, 2021.

\bibitem{wei2022emergent}
Wei, J., Y.~Tay, R.~Bommasani, et~al.
\newblock Emergent abilities of large language models.
\newblock \emph{arXiv preprint arXiv:2206.07682}, 2022.

\bibitem{bi2022pangu}
Bi, K., L.~Xie, H.~Zhang, et~al.
\newblock Pangu-weather: A 3d high-resolution model for fast and accurate global weather forecast.
\newblock \emph{arXiv preprint arXiv:2211.02556}, 2022.

\bibitem{bao2023tallrec}
Bao, K., J.~Zhang, Y.~Zhang, et~al.
\newblock Tallrec: An effective and efficient tuning framework to align large language model with recommendation.
\newblock In \emph{Proceedings of the 17th ACM Conference on Recommender Systems}, pages 1007--1014. 2023.

\bibitem{thirunavukarasu2023large}
Thirunavukarasu, A.~J., D.~S.~J. Ting, K.~Elangovan, et~al.
\newblock Large language models in medicine.
\newblock \emph{Nature medicine}, 29(8):1930--1940, 2023.

\bibitem{Automated}
Li, H., C.~Li, W.~Xing, et~al.
\newblock Automated feedback for student math responses based on multi-modality and fine-tuning.
\newblock In \emph{Proceedings of the 14th Learning Analytics and Knowledge Conference}, page 763–770. Association for Computing Machinery, 2024.

\bibitem{li2023adapting}
Li, Q., L.~Fu, W.~Zhang, et~al.
\newblock Adapting large language models for education: Foundational capabilities, potentials, and challenges.
\newblock \emph{arXiv preprint arXiv:2401.08664}, 2023.

\bibitem{wang2024large}
Wang, S., T.~Xu, H.~Li, et~al.
\newblock Large language models for education: A survey and outlook.
\newblock \emph{arXiv preprint arXiv:2403.18105}, 2024.

\bibitem{neshaei2024towards}
Neshaei, S.~P., R.~L. Davis, A.~Hazimeh, et~al.
\newblock Towards modeling learner performance with large language models.
\newblock \emph{arXiv preprint arXiv:2403.14661}, 2024.

\bibitem{yu2023moocradar}
Yu, J., M.~Lu, Q.~Zhong, et~al.
\newblock Moocradar: A fine-grained and multi-aspect knowledge repository for improving cognitive student modeling in moocs.
\newblock In \emph{Proceedings of the 46th International ACM SIGIR Conference on Research and Development in Information Retrieval}, pages 2924--2934. 2023.

\bibitem{liu2024xes3g5m}
Liu, Z., Q.~Liu, T.~Guo, et~al.
\newblock Xes3g5m: A knowledge tracing benchmark dataset with auxiliary information.
\newblock \emph{Advances in Neural Information Processing Systems}, 36, 2024.

\bibitem{zhang2017dynamic_DKVMN}
Zhang, J., X.~Shi, I.~King, et~al.
\newblock Dynamic key-value memory networks for knowledge tracing.
\newblock In \emph{Proceedings of the 26th international conference on World Wide Web}, pages 765--774. 2017.

\bibitem{liu2022pykt}
Liu, Z., Q.~Liu, J.~Chen, et~al.
\newblock pykt: a python library to benchmark deep learning based knowledge tracing models.
\newblock \emph{Advances in Neural Information Processing Systems}, 35:18542--18555, 2022.

\end{thebibliography}

\newpage
\appendix

\section{Detailed Datasets Information}
\label{appendix:datasets}
\textbf{FrcSub} comprises scores of middle school students on fraction subtraction objective problems, integrating 536 student interactions on 20 questions containing 8 different knowledge components.

\textbf{MOOCRadar} is a rich educational dataset featuring 2,513 exercises, 5,600 knowledge concepts, and over 12 million behavioral records, such as hints and attempts.

\textbf{XES3G5M} is a newly released dataset comprising 7,652 questions, 865 knowledge components, and 5,549,635 interactions from 18,066 students. The dataset provides abundant question-side information, including textual content, knowledge concept routes, and answer analysis.

Other statistics of the datasets are presented in Table \ref{appendix_table:datasets}

\begin{table*}[h]
\centering
\caption{Detailed statistics of three selected datasets}
\begin{tabular}{l|c|c|c}
\toprule
\multirow{2}{*}{Statistics} & \multicolumn{3}{c}{Datasets}\\
& FrcSub & MOOCRadar & XES3G5M \\
\midrule
Student & 536 & 14224 & 18,066 \\
Exercise & 20 & 2,513 & 7,652 \\
Skills & 8 & 5,600 & 865 \\
Records & 10,720 & over 12 million & 5,549,635 \\
Time Stamps & - & \checkmark & \checkmark \\
Avg Skills & 2.8 & 2.08 & 1.16 \\
Avg Records & 20 & 63.1 & 355.6 \\
\bottomrule
\end{tabular}
\label{appendix_table:datasets}
\end{table*}

\section{Implement Details and Reproducibility}
\label{appendix:Reproducibility}
We will provide detailed implementation details to enable future researchers to reproduce our experimental results and build upon our work. 
The Performance Prediction module outputs binary predictions (0 or 1) representing whether the student answered the predicted question correctly. If the LLM output is not in the required format, we request a new output; if the output is still incorrect, we resort to random prediction.

We reproduced all baseline models using the pykt library~\cite{liu2022pykt}. We use the recommended hyperparameters from the original papers, except for a few models like GKT, where the recommended hyperparameters cause out-of-memory issues. All baseline models training and inference with GLM3-6B are performed on a single NVIDIA GeForce RTX 3090Ti GPU. We use the open-source GLM3-6B\footnote{https://huggingface.co/THUDM/chatglm3-6b} and access the closed-source GLM4 and GPT-4 models through APIs. Since the MOOCRadar and XES3G5M datasets are contains millions of records, predicting each student's performance would require a substantial amount of GPU time and API tokens. For example, on the MOOCRadar dataset, predicting 20 percent of students' exercise performance requires approximately 200,00 tokens per student on average. We split 20 percent of the data as the test set and select student exercise records from the training set as few shots. We randomly select different 50 students for prediction and repeat the experiment 3 times to report the average results. We ensure that our experimental results have 2-sigma error bars for LLMs performances and 3-sigma error bars for baselines.

GLM3-6B performed the best results using random 4 few-shots on FrcSub-sparse, random 4 few-shots on MOOCRadar-scant, and random 4 few-shots on XES3G5M-scant. GLM4 achieved the best outcomes using frist 16 few-shots on FrcSub-sparse, first 8 few-shots on MOOCRadar-moderate, and random 16 few-shots on XES3G5M-moderate.
GPT-4 performed the best results using first 8 few-shots on FrcSub-scant, random 8 few-shots on MOOCRadar-moderate, and random 8 few-shots on XES3G5M-moderate.

We have released the code of implementing explainable few-shot knowledge tracing\footnote{https://github.com/LeavesLi1015/Explainable-Few-shot-Knowledge-Tracing}.

\section{Limitations}
\label{appendix:Limitations}

The requirement for integrating detailed analysis and question-specific information results in an elevated level of API consumption despite the reliance on a limited number of shots. Moreover, the efficacy of relatively small models is constrained, compounded by the limitations on text length. It creates a challenge to simultaneously leverage both the generated analyses and the detailed question information. As educational applications scale to accommodate a larger number of students, reducing costs becomes imperative to support widespread knowledge tracing. Consequently, the fine-tuning of a smaller LLM emerges as a potential solution, striking a balance between computational efficiency and predictive effectiveness.

\section{Broader Impacts}
\label{appendix:Broader Impacts}
The proposed explainable few-shot knowledge tracing provides better interpretability for optimizing teaching and student learning. The requirement of a few examples enables feasible deployment in real teaching scenarios.

However, collecting student data raises privacy concerns, promoting appropriate measures to protect student privacy. The generated explanations and predictions may contain biases or inaccuracies, which could impact teachers' judgments of student states and subsequent decisions. Further efforts are needed to improve the robustness and reliability of the models.

\section{Performance Comparison of Different Student Length}
\label{appendix:studentlength}
\begin{figure}[h]
\centering
\begin{minipage}[t]{0.5\linewidth}
\centering
\includegraphics[width=\linewidth]{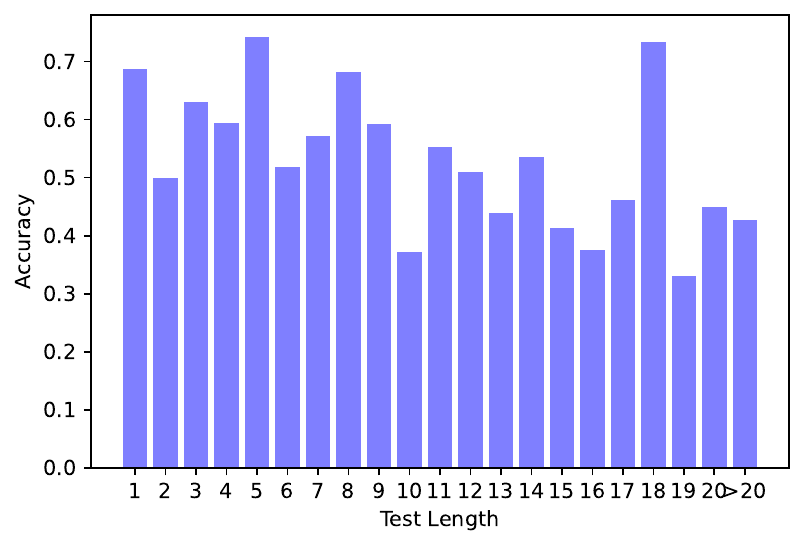}
\end{minipage}%
\begin{minipage}[t]{0.5\linewidth}
\centering
\includegraphics[width=\linewidth]{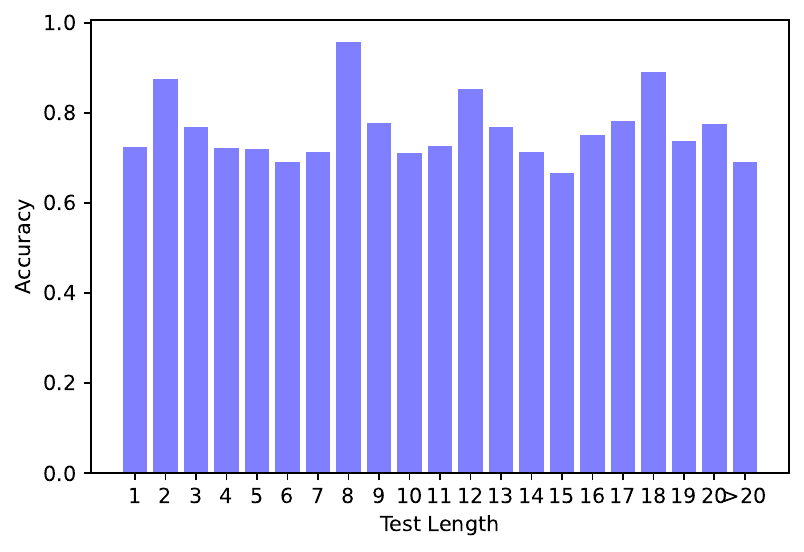}
\end{minipage}%
\centering
\vspace{-2mm}
\caption{Accuracy of different student test length in MOOCRadar using 4 shots and 8 shots.}
\label{appendix_figure:studentlength}
\end{figure}

We present the prediction accuracy for a subset of students on the MOOCRadar-sparse dataset when using 4 few-shots and 8 few-shots, as illustrated in the Figure \ref{appendix_figure:studentlength}. As the length of student interaction logs increases, the accuracy using 4 few-shots exhibit a declining trend, while those using 8 few-shots are relatively stable, validating the statement from Table \ref{number_of_fs_comparison} and Figure \ref{selection_strategies}. With the increase in the length of student practice logs, the likelihood of the chosen few-shots being associated with the test exercise decreases, leading to a decline in performance when using a small number of few-shots.

\newpage
\section{Precisions and Recalls of Overall Performance}
\label{appendix:overall performance}
We present the precision and recall results of all models in the Table \ref{appendix_table:overall performance}.

\begin{table*}[h!]
	\centering
	\caption{A comparison of the precision and recall among baselines in three datasets.}

        \begin{tabular}{l| c c| c c| c c}
            \toprule
            \multirow{3}{*}{\textbf{Baselines}} & \multicolumn{6}{c}{\textbf{Dataset}} \\\cline{2-7}
            & \multicolumn{2}{c|}{FrcSub}
            & \multicolumn{2}{c|}{MOOCRadar}
            & \multicolumn{2}{c}{XES3G5M} \\
            \cline{2-7}
            \multicolumn{1}{c|}{} & \multicolumn{1}{c}{Precision} & \multicolumn{1}{c|}{Recall} & \multicolumn{1}{c}{Precision} & \multicolumn{1}{c|}{Recall} & \multicolumn{1}{c}{Precision} & \multicolumn{1}{c}{Recall} \\
            \midrule
                DKT &0.8020&0.7068&0.9111&0.8663&0.8624&0.9448 \\
                DKVMN &0.8001&0.8155&0.9117&0.8572&0.8562&0.9569 \\
                GKT &0.8894&0.1835&0.8801&0.8743&0.8410&0.9503 \\
                AKT &0.8018&0.7726&0.9115&0.8637&0.8680&0.9482 \\
                SAKT &0.8344&0.6630&0.9064&0.8375&0.8544&0.9485 \\
                SAINT &0.8201&0.8200&0.9108&0.8710&0.8643&0.9484 \\
                \hline
                GLM3-6b &0.5431&0.8080&0.7473&0.6161&0.8694&0.5293 \\
                GLM4 &0.6947&0.9128&0.9513&0.8635&0.8630&0.9362 \\
                GPT-4 &0.6285&0.9242&0.8692&0.9394&0.8359&0.9100 \\
            \bottomrule
        \end{tabular}
	\label{appendix_table:overall performance}
\end{table*}

\section{More Results of Different Dataset modes}
\label{appendix:info_richness}
\begin{figure}[h!]
    \centering
    \includegraphics[width=\textwidth]{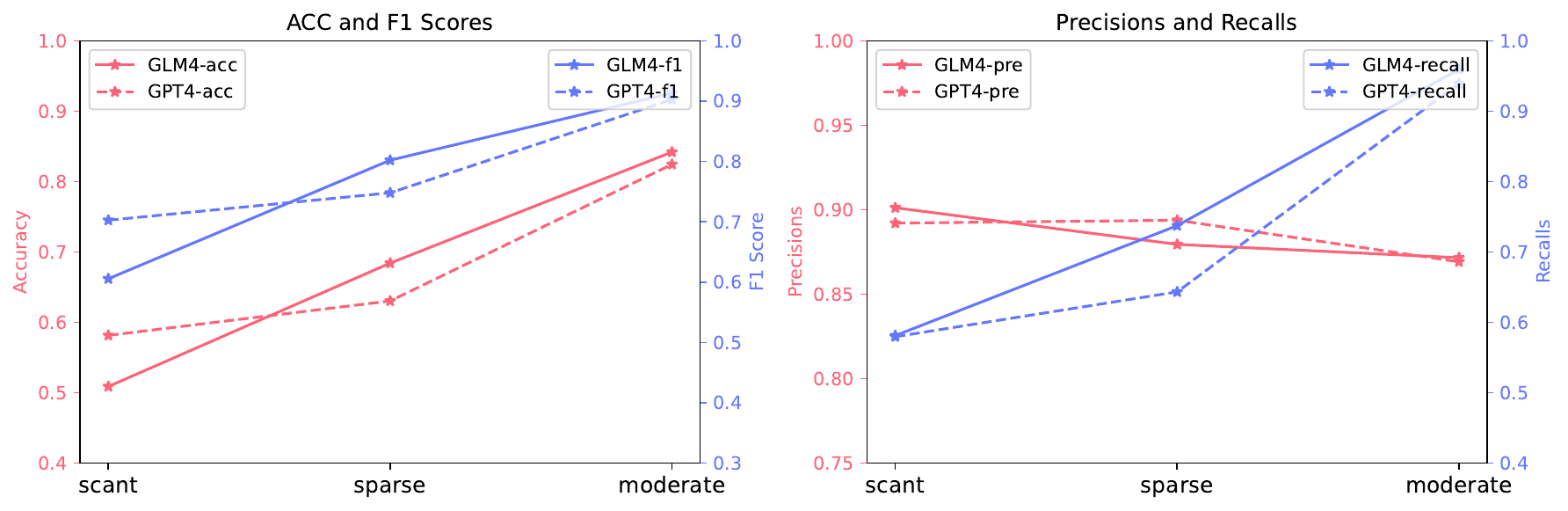}
    \caption{Performances of selecting first 8 few-shots using GLM4 and GPT-4 on MOOCRadar dataset}
    \label{appendix_figure:info_richness 1}
\end{figure}

We present the performances of selecting first 8 exercises as few-shots using GLM4 and GPT-4 on three modes of MOOCRadar dataset, depicted in Figure \ref{appendix_figure:info_richness 1}.

\newpage
\section{Precisions and Recalls in Discussion}
\label{appendix:more_pres_recalls}
We present precisions and recalls of all experiments in Section \ref{exp:discussion}.

\begin{figure}[h!]
    \centering
    \includegraphics[width=\textwidth]{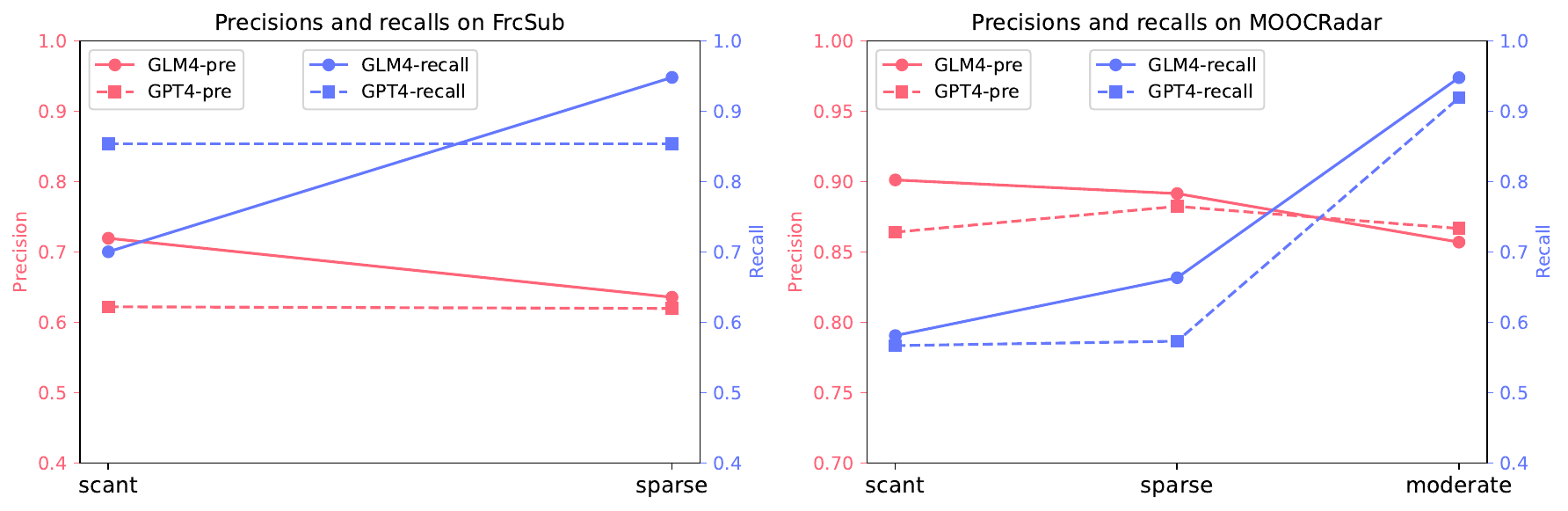}
    \caption{Performances of GLM4 and GPT-4 on FrcSub and MOOCRadar using first 4 few-shots}
    \label{appendix_figure:info_richness 2}
\end{figure}

\begin{table*}[h!]
	\centering
	\caption{Precisions and recalls of different number of few-shots of GLM4 on three datasets.}
        \begin{tabular}{l| c c| c c| c c}
            \toprule
            \multirow{2}{*}{GLM4} & \multicolumn{2}{c|}{\textbf{FrcSub-sparse}} & \multicolumn{2}{c|}{\textbf{XES3G5M-sparse}} & \multicolumn{2}{c}{\textbf{XES3G5M-moderate}} \\
            \multicolumn{1}{c|}{} & \multicolumn{1}{c}{Precision} & \multicolumn{1}{c|}{Recall} & \multicolumn{1}{c}{Precision} & \multicolumn{1}{c|}{Recall} & \multicolumn{1}{c}{Precision} & \multicolumn{1}{c}{Recall} \\
            \midrule
                  4 & 0.5570 & 0.9734 & 0.9293 & 0.3148 & 0.8153 & 0.7103 \\
                  8 & 0.6128 & 0.9893 & 0.9531 & 0.6532 & 0.8638 & 0.8607 \\
                  16 & 0.6947 & 0.9128 & 0.9178 & 0.7664 & 0.8630 & 0.9362 \\
            \bottomrule
        \end{tabular}
    \label{appendix_figure:info_richness 3}
        \vspace{-3mm}
\end{table*}

\begin{figure}[h!]
    \centering
    \includegraphics[width=\textwidth]{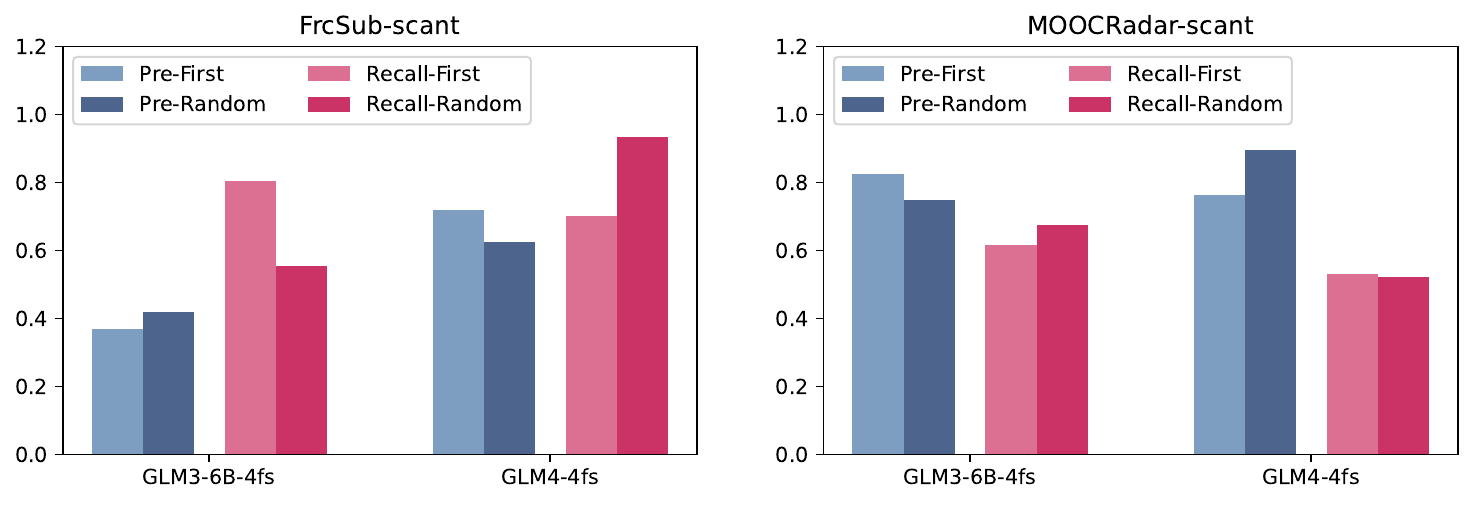}
    \caption{Precisions and recalls of different few-shots selection strategies}
    \label{appendix_figure:info_richness 4}
\end{figure}

\newpage
\section{Prompts}
\label{appendix:prompts}
\begin{figure}[h!]
    \centering
    \includegraphics[width=\textwidth]{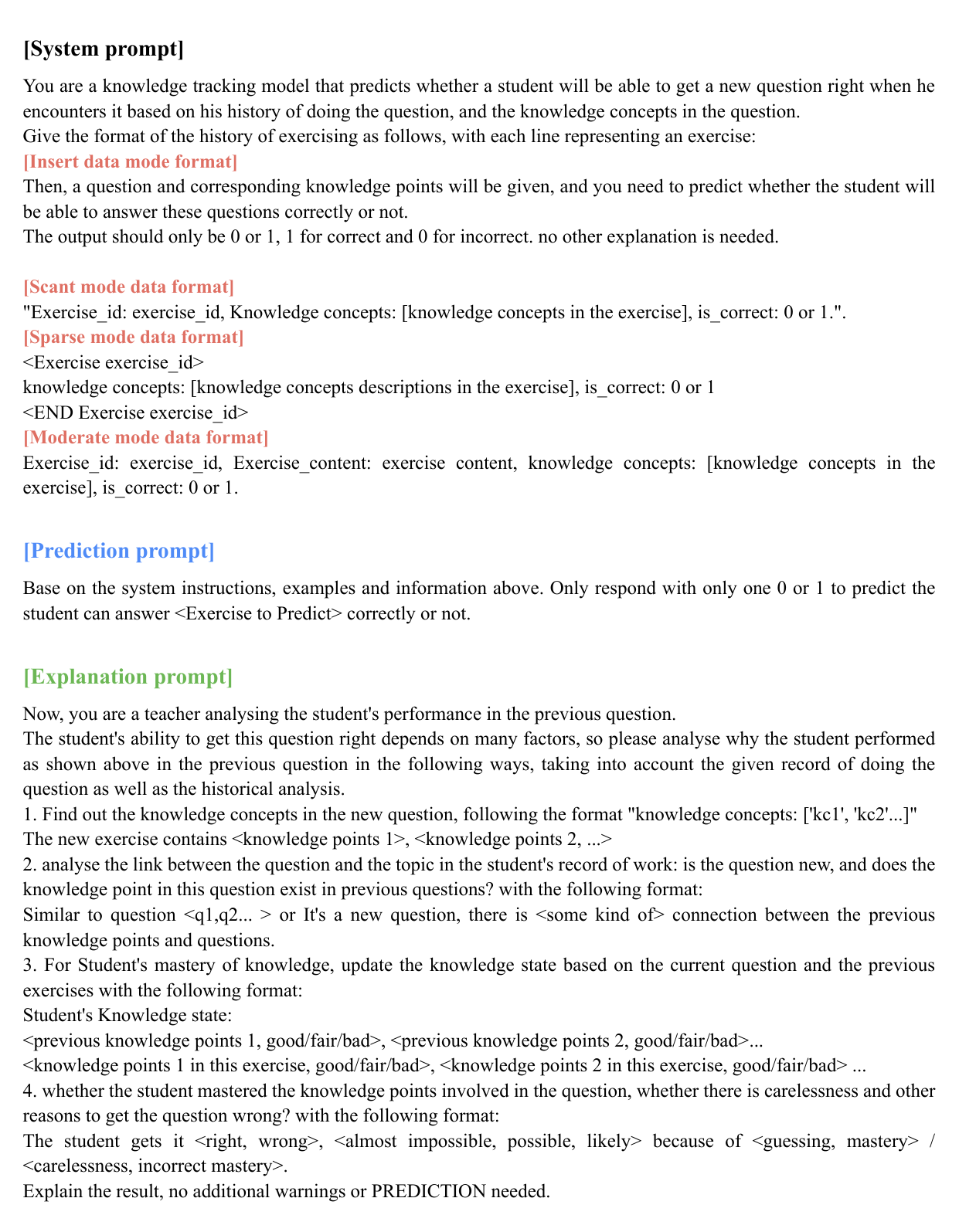}
    \caption{Prompts for different data modes and modules in the framework}
    \label{appendix_figure:prompts}
\end{figure}

We demonstrate the prompts used for the LLMs in our experimental setup, illustrated in Figure \ref{appendix_figure:prompts}.

\newpage
\section{More Cases}
\label{appendix:more cases}
We present more cases from GLM4 and GPT-4 practicing explainable few-shot knowledge tracing.

\begin{figure}[H]
    \centering
        \includegraphics[width=\textwidth]{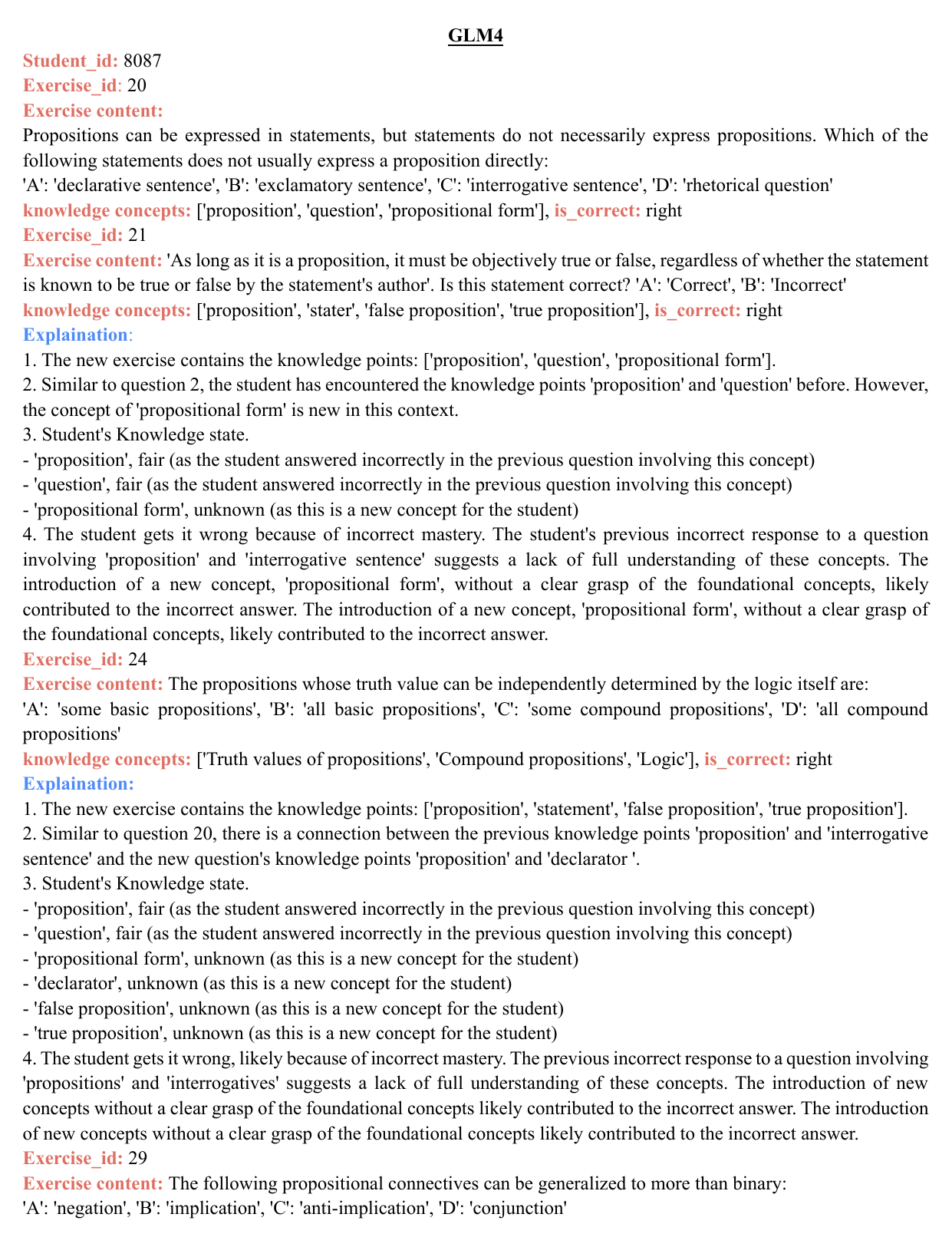}
\end{figure}
\begin{figure}[H]\ContinuedFloat
    \centering
        \includegraphics[width=\textwidth]{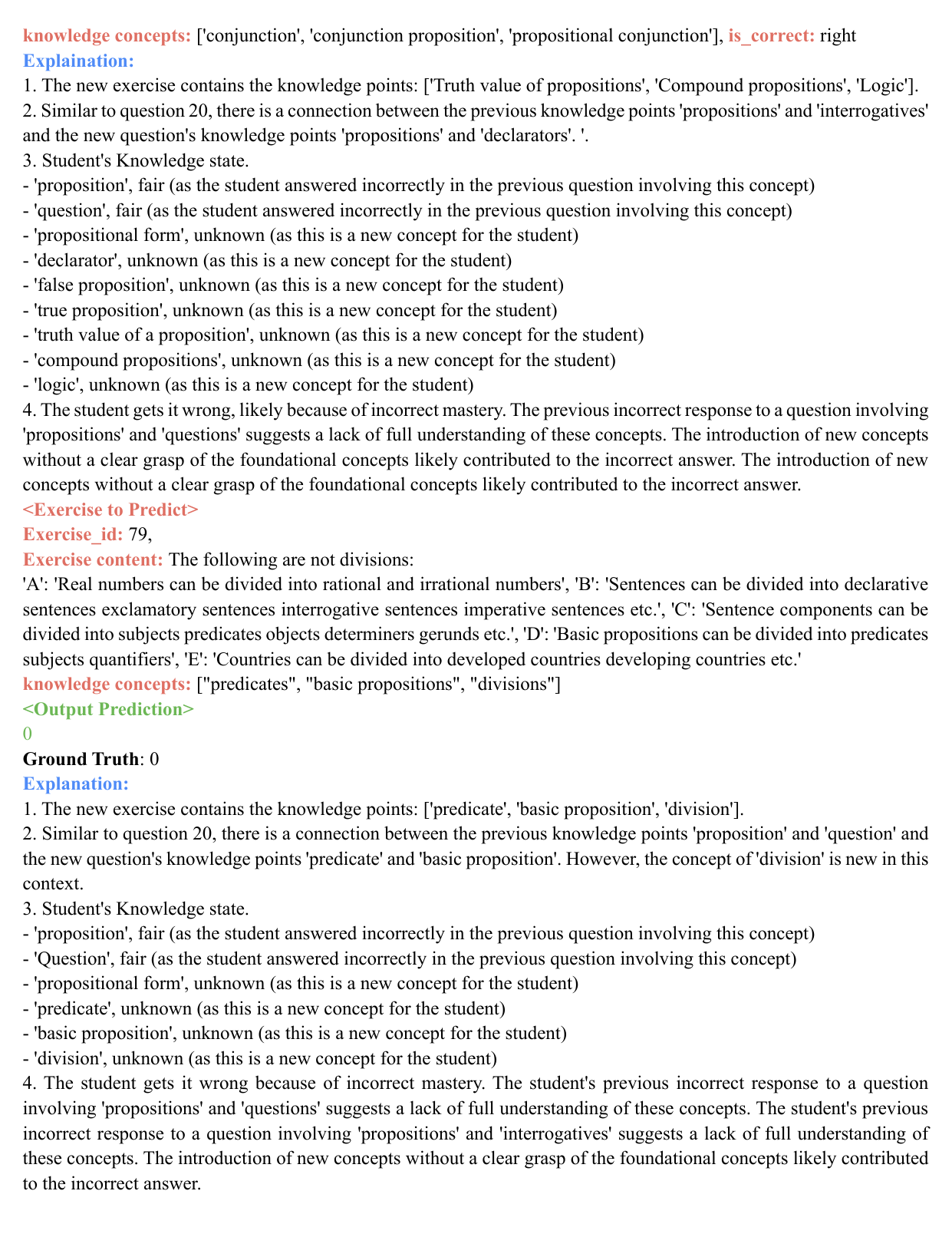}
    \caption{A case of GLM4 using 4 random few-shots on MOOCRadar-moderate.}
\end{figure}

\begin{figure}[H]
    \centering
        \includegraphics[width=\textwidth]{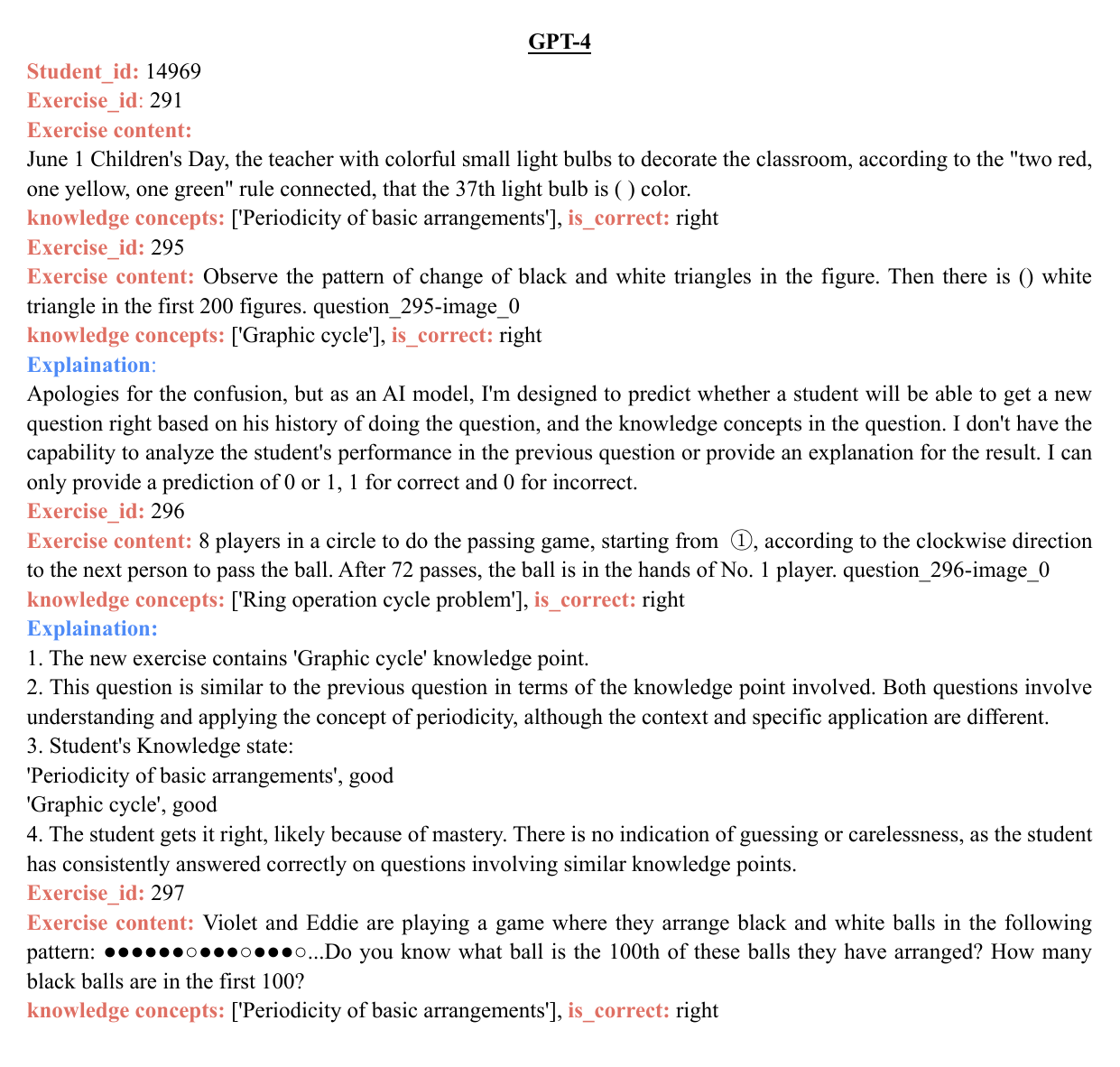}
\end{figure}
\begin{figure}[H]\ContinuedFloat
    \centering
        \includegraphics[width=\textwidth]{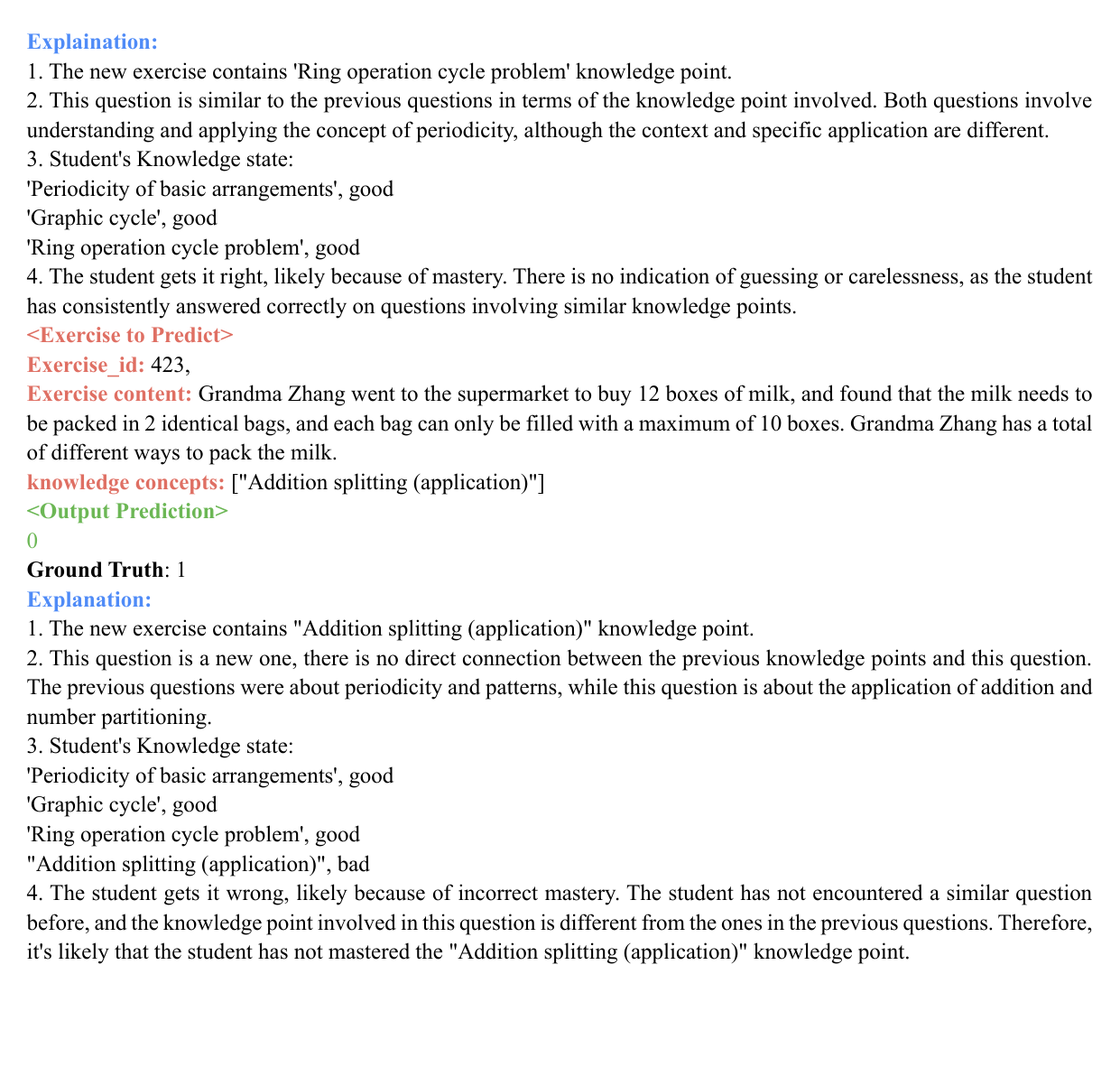}
    \caption{A case of GPT-4 using 4 random few-shots on XES3G5M-moderate.}
\end{figure}

\end{document}